\documentclass{svjour3}

\usepackage{natbib}
\usepackage[utf8]{inputenc}
\usepackage{enumerate}
\usepackage[dvipsnames]{xcolor}
\usepackage{eurosym}
\usepackage{amsmath}
\usepackage{amssymb}
\usepackage{amsfonts}
\usepackage{graphicx}
\usepackage{hyperref}

\usepackage[noabbrev]{cleveref}
\usepackage{tikz}
\usepackage{geometry}
\usepackage{floatpag}
\usepackage{mathrsfs}
\usepackage{mathbbol}
\usepackage{mathbbol}
\usepackage{mathrsfs}
\usepackage{stmaryrd}
\usepackage{txfonts}

\usetikzlibrary{positioning,arrows,decorations.markings}
\usetikzlibrary{decorations.pathreplacing,angles,quotes}
\tikzset{
  node/.style={circle,draw,minimum size=8ex},
  arrow/.style={-latex,draw,minimum size=10ex}}
\spnewtheorem*{defn}{Definition}{\bf}{\rm}

\newcommand{\N}{\mathbb{N}}
\newcommand{\R}{\mathbb{R}}

\hypersetup{
    colorlinks=true,
    linkcolor=blue,
    filecolor=magenta,      
    urlcolor=cyan,
    citecolor=blue
}
\def\makeheadbox{\relax}

\begin{document}
\title{The Intriguing Relation Between Counterfactual Explanations and Adversarial Examples}
\author{Timo Freiesleben}

\institute {Timo Freiesleben\\
Ludwigstrasse 31\\
Munich Center for Mathematical Philosophy\\ Ludwig-Maximilians-Universität\\ Munich\\
Germany\\
ORCiD: 0000-0003-1338-3293
\\\email{Timo.Freiesleben@campus.lmu.de}
}

\date{}

\maketitle

\begin{abstract}
The same method that creates adversarial examples (AEs) to fool image-classifiers can be used to generate counterfactual explanations (CEs) that explain algorithmic decisions. This observation has led researchers to consider CEs as AEs by another name. We argue that the relationship to the true label and the tolerance with respect to proximity are two properties that formally distinguish CEs and AEs. Based on these arguments, we introduce CEs, AEs, and related concepts mathematically in a common framework. Furthermore, we show connections between current methods for generating CEs and AEs, and estimate that the fields will merge more and more as the number of common use-cases grows.
\end{abstract}

\begin{acknowledgements}
This work was supported by the Graduate School of Systemic Neuroscience (GSN) of the LMU Munich.
We want to thank Stephan Hartmann, Christoph Molnar, Gunnar König, and the GSN Neurophilosophy-group for their helpful comments on the manuscript, the fruitful discussions about the concepts, and their hints to related literature. Also big thanks to all anonymous reviewers that gave helpful feedback on the paper.
\end{acknowledgements}

\newpage

\section{Introduction}
\label{sec:intro}
Machine Learning (ML) is transforming industry, science, and our society. Today, ML algorithms can fix a date at the hairdresser \citep{duplex}, determine a protein’s 3D shape from its amino-acid sequence \citep{senior2020improved}, and even write news articles \citep{brown2020language}. Taking a sharp look at these developments, we observe a tendency towards more and more complex models. Different ML models are stacked together heuristically, with limited theoretical backing \citep{hutson2018ai}. In some applications, complexity may not be an issue as long as the algorithm performs well most of the time. However, in socially, epistemically, or safety-critical domains, complexity can rule out ML solutions - think of e.g. autonomous driving, scientific discovery, or criminal justice. Two of the major drawbacks of highly complex algorithms are the \emph{opaqueness problem} \citep{lipton2018mythos}  and \emph{adversarial attacks} \citep{szegedy}.

The opaqueness problem describes the limited epistemic access humans have to the inner workings of ML algorithms, especially concerning the semantic interpretation of parameters, the learning process, and the human-predictability of ML decisions \citep{burelldoi:10.1177/2053951715622512}.
This lack of interpretability has gained a lot of attention recently, which gave rise to the field eXplainable Artificial Intelligence (XAI) \citep{doshi2017towards,rudin2019stop}. Many techniques have been proposed to gain insights into ML systems \citep{adadi2018peeking,dovsilovic2018explainable,das2020opportunities}. Especially model-agnostic methods have gained attraction since, unlike model-specific methods, their application is not restricted to a specific model type \citep{molnar2019}. Global model-agnostic interpretation techniques like Permutation Feature Importance \citep{fisher2019all} or Partial Dependence Plots \citep{friedman1991multivariate} aim at understanding the general properties of ML algorithms. On the other side, local model-agnostic interpretation methods like LIME \citep{ribeiro2016should} or Shapley Values \citep{vstrumbelj2014explaining} aim at understanding the behaviour of algorithms for particular regions. One way to explain a specific model-prediction is a Counterfactual Explanation (CE) \citep{wachter2017counterfactual}. A CE explains a prediction by presenting a maximally close alternative input that would have resulted in a different (usually desired) prediction. CEs are the first class of objects, we study in the present paper.

The problem of adversarial attacks describes the fact that complex ML algorithms are vulnerable to deceptions \citep{papernot2016transferability,goodfellowEx,szegedy}. Such malfunctions can be exploited by attackers to e.g. harm model-employers or endanger end-users \citep{song2018physical}. The field that investigates adversarial attacks is called adversarial machine learning \citep{joseph2018adversarial}. If the attack happens during the training process by inserting mislabeled training data the attack is called poisoning. If an attack happens after the training process it is commonly called an adversarial example (AE) \citep{serban2020adversarial}. AEs are inputs that resemble real data but are misclassified by a trained ML model, e.g., the image of a turtle is classified as a riffle \citep{athalye2018synthesizing}. Hence, misclassified means here that the algorithm assigns the wrong class/value compared to some (usually human-given) ground-truth \citep{elsayed2018adversarial}. AEs are the second class of objects relevant to our study.

Even though the opaqueness problem and the problem of adversarial attacks seem unrelated at first sight, there are good reasons to study them jointly. AEs show where an ML model fails, and examining these failures deepens our understanding of the model \citep{tomsett2018failure,dong2017towards}. Explanations on the other hand can shed light on how ML algorithms can be improved to make them more robust against AEs
\citep{molnar2019}. As a downside, explanations may enclose too much information about the model, thereby allowing AEs to be constructed and the model attacked
\citep{ignatiev2019relating,SokolFlach}. CEs are even stronger connected with AEs than other explanations. CEs and AEs can be obtained by solving the same optimization problem\footnote{$x$ describes the original input, $x'$ the counterfactual/adversarial vector, $f$ the ML model, $y_{des}$ the desired classification, $d(\cdot,\cdot)$ and $d'(\cdot,\cdot)$ distances, and $\lambda$ a trade-off scalar. For details, see \Cref{subsec:definitions}.} \citep{wachter2017counterfactual,szegedy}:
\begin{equation}
    \label{eq:optimization}
    \underset{x'\in X}{\text{argmin}}\; d(x,x') +\lambda\; d'(f(x'),y_{des}) 
\end{equation}
 Term \ref{eq:optimization} has led to various confusions concerning the relationship between CEs and AEs in the research community.\footnote{We discuss these confusions in more detail in \Cref{subsec:conDis}.} We aim to resolve them and give a detailed analysis of the relationship between the two fields.

The aim of the present paper is twofold. Our first goal is the \emph{clarification of concepts}. Commonly used concepts such as CE/AE, flipping/misclassifying, process/model-level, and closeness/distance are often misunderstood or not clearly defined. We define these terms properly in one mathematical framework, aiming for more clarity and unification. The second goal is to \emph{familiarize researchers} of each of the respective fields \emph{with its neighboring area}. Even in one of the fields, it is hard to keep track of developments and new ideas, in both it is worse. Since there are many ways in which each of the fields can profit from the other, both methodologically and conceptually, we aim to provide a guide connecting the two literatures.

We will start by providing an intuition to the reader with two standard use cases of CEs/AEs and give an overview of relevant other applications in \Cref{sec:useCases}. In \Cref{sec:background}, we present the (historical) background of CEs and AEs, including the current debate around their relationship. Next, we present arguments in what sense the current understanding of the relation between CEs and AEs is flawed in \Cref{subsec:conDis}. We will argue that the notions of misclassification and maximal proximity are the central properties that distinguish CEs from AEs. Based on that, we introduce in \Cref{subsec:definitions} our more fine-grained formal definitions of CEs, AEs, and related concepts. In \Cref{sec:Generation}, we discuss connections between the solution approaches for finding CEs/AEs in the literature. We conclude in \Cref{sec:discussion} by discussing the relevance and limitations of our work.

\section{Examples and Use Cases}
\label{sec:useCases}
Before we get into the technical and conceptual details, let us look at two use cases where both CEs and AEs have been successfully deployed. This provides an intuition to the reader and will moreover serve for explanatory purposes in the later sections. The first example is among the most prominent use-cases of CEs, automated lending. The second example shows one prominent use-case of AEs, image-classification of hand-written digits.

\paragraph{Loan Application:}Imagine a scenario where person P wants to obtain a loan and applies for it through a bank's online portal. She has to enter several of her properties into the user-interface e.g. her age, salary, capital, number of open loans, and number of pets. The portal uses an automated, algorithmic decision system, which decides that P will not receive the loan. However, she would have liked to obtain it and therefore demands an explanation. An example of a potential CE would be: 
\begin{itemize}
    \item[] If P had a $5,000$ \euro\, p.a. higher salary and an outstanding loan less, her loan application would have been accepted.
\end{itemize}
She can use this information to guide her future actions or potentially to contest the algorithmic decision. Clearly, CEs are not restricted to that setting. If P were the model engineer instead of the customer, she could also use the explanation to raise her understanding of the model or to debug it.

Now, suppose that P wants to trick the system to get the credit. One potential way to trick the system with an AE could for example look as follows:
\begin{itemize}
    \item[] P indicates two more pets on the application form than she actually has to obtain the loan.
\end{itemize}
P has changed a feature that she probably does not have to prove to the bank. This change allowed her to obtain the loan, even though none of her properties have changed.

\paragraph{Hand-Written Digits Recognition:}
Imagine a scenario in which a postal service employs an image recognition algorithm. This algorithm takes as input grey-scale $28\times28$ pixel images and assigns them the number between 0-9 they depict. This procedure eases the work of the postal service a lot. Cases of errors are rare but costly, as the postal service must pay the sender $5$\euro\, if a letter or package is sent to the wrong address. Therefore, the postal service is interested in improving the algorithm.

\begin{figure}[h]
\centering
  \includegraphics[width=0.3\linewidth]{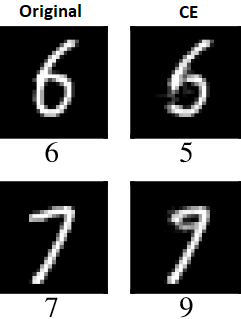}\qquad \includegraphics[width=0.33\linewidth]{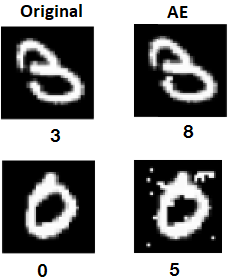}
  \caption{The images are taken from \cite{van2019interpretable} and \cite{papernot2017practical}. They are generated from CNNs trained on the MNIST dataset. The first an the third column depict original images from the MNIST dataset. Column two depicts the corresponding CEs and column four shows the corresponding AEs.}
  \label{fig:mnist}
\end{figure}

One option would be to generate CEs for specific instances, evaluate how useful they are, and adjust the algorithm. Such CEs can be found in the first two columns of \Cref{fig:mnist}. One can see e.g. that the images in the first row show that the algorithm assigns major importance to the lower-left line to distinguish between a six and a five. The postal service might derive that the algorithm already has a robust understanding of digits.

Now, assume we take the perspective of an attacker who is interested in exploiting the $5$\euro\, per error system. Such an attacker will be interested in generating AEs, put them on letters/parcels and gain money. Examples of such AEs are presented in the last column of \Cref{fig:mnist}. One can see e.g. that the algorithm has problems when random dots appear around a $0$ and misclassifies the input as the number $5$. While the attacker will be happy to accomplish many successful attacks, the postal service will try to limit the deceivability of its algorithm to save money.

\paragraph{The Relevance of These Use-Cases:}
Loan applications are among the most popular example use-cases in the CE literature \citep{wachter2017counterfactual,dandlmulti,grath2018interpretable}. The example is particularly valuable as it describes a technically and ethically complex decision situation, in which explanations are a requirement. Interestingly, the lending use-case gains more and more interest also in the AE literature since it depicts the safety troubles of ML systems. \cite{ballet2019imperceptible} introduce a new notion of imperceptibility for these scenarios which got quickly picked up by others \citep{cartella2021adversarial,hashemi2020permuteattack}.

Hand-Written-Digits classification is the classical use-case among all image-classification tasks. Many methods to generate AEs discuss it at least as a test case \citep{wang2019gan,szegedy,papernot2017practical}. The feature space is comparatively small and the problem itself well studied, therefore, generating AEs is computationally cheap and conceptually informative. However, security threats cannot be as easily depicted from this use case. Because of its simplicity, it has also been used as a starting point in the CE literature. The difficulty lies here in finding semantically meaningful notions of similarity for images. Three papers proposed approaches to that problem,  \cite{van2019interpretable} use prototypes to generate realistic CEs, \cite{poyiadzi2020face} use allowed paths, and \cite{goyal2019} use differently classified images to identify regions that shift the classification.

\paragraph{Other Use Cases:}
There are common use-cases for CEs other than loan approval, such as university applications, diabetes diagnosis \citep{wachter2017counterfactual}, adult-income prediction \citep{mothilal2019explaining}, or predicting student performances in law-school \citep{RussellCoherent}. Most of the common use-cases focus on tabular data settings, as it is easier to make sense of CEs in these scenarios \citep{verma2020counterfactual}. Changes in semantically meaningful variables are easy to convey. Moreover, the scenarios considered often describe high-stakes decisions with an ethical dimension. There are few non-classification, non-tabular settings in which CEs have been applied, such as image recognition  \citep{goyal2019,van2019interpretable}, NLP-tasks \citep{akula2019natural}, regression problems \citep{sule2019} and non-supervised learning settings \citep{olson2021counterfactual}.

The AE-community on the other hand is largely focused on image classification tasks \citep{serban2020adversarial}. Many AEs focus particularly on the state-of-the-art image classifiers from Google, Amazon, or Facebook \citep{serban2020adversarial}. Well-known examples include AEs on road signs \citep{eykholt2018robust}, the 3-D print of a turtle classified as a rifle \citep{athalye2018synthesizing}, and the adversarial patch, a sticker that fools image recognition software into classifying it as a toaster \citep{brown2017adversarial}. One reason why image classifiers are mostly focused on is that the imperceptibility of changes and the true class label are easy to define \citep{ballet2019imperceptible}. Moreover, since image recognition models focus on models like CNNs, AEs help assessing the limitations of opaque deep learning algorithms. However, there is also work on AEs in other tasks e.g. audio/video-classification \citep{carlini2016hidden,carlini2018audio,wei2018transferable}, regression problems \citep{balda2019perturbation}, and non-supervised learning settings \citep{behzadan2017vulnerability,huang2017adversarial}.

\section{Background on CEs and AEs}
\label{sec:background}
This section provides a background on where CEs and AEs have historically come from, discusses their roles in ML, and presents the discussions about the relationship between the two. The historic background and roles of CEs/AEs provide the basis for understanding the discussions around the relationship between the two fields, which motivate our proposal.

\subsection{Historic Background}
\label{subsec:history}
\paragraph{History of CEs:}CEs have their roots in Philosophy as so-called \emph{subjunctive counterfactual conditionals}. They describe conditionals of the form    
\begin{equation}
    \label{eq:counterfactual}
        \text{If }S\text{ was the case }Q\text{ would have been the case.}
    \end{equation} 
where $S$ and $Q$ are events. Importantly, event $S$ did not in fact occur. The truth-condition for conditional \ref{eq:counterfactual} is hotly debated in philosophy until today \citep{sep-counterfactuals}.
The approach taken up by the XAI community \citep{wachter2017counterfactual} builds on the work of \cite{Lewis1973} and \cite{stalnaker1968theory}. In their framework, conditional \ref{eq:counterfactual} holds if and only if the closest possible world\footnote{$\Omega$ denotes the set of possible worlds.} $\omega'\in\Omega$ to the actual world $\omega\in\Omega$ in which $S$ is the case\footnote{As mentioned, above $S$ is false in $\omega$.} also $Q$ is the case. The notion of similarity between possible worlds is critical in assessing a counterfactual conditional and Lewis discusses similarity in more detail in \cite{lewis1979counterfactual}. He argues that between close worlds laws of nature must be preserved, widespread, diverse violations should be avoided, and facts stay congruent for maximal time. Particular facts on the other side can be changed without significantly increasing dissimilarity. Despite these specifications, Lewis himself admits that the under-specified notion of similarity between possible worlds remains the crucial weak-spot of his framework \citep{lewis1983philosophical}.

It is very important to keep in mind that Lewis aimed to describe causal dependence via counterfactual conditionals \citep{sep-causation-counterfactual}. The idea is that $Q'$ causally depends on $S'$ if and only if, if $S$ were not the case $Q$ would not have been the case.\footnote{Interestingly, \cite{pearl2009causality} turns this story around and defines counterfactuals via causal graphs. Instead of comparing similar worlds, he directly focuses on the underlying mechanisms defined by a structural equation. However, as \cite{woodward2002mechanism} and \cite{hitchcock2001intransitivity} pointed out that is a matter of interpretation as we can instead also understand Pearl's structural equations as sets of primitive counterfactuals. Also Pearl's notion has found its way into the XAI literature in the form of algorithmic recourse \citep{KarSchVal20,karimi2020survey}.} Even though CEs must not be causal \citep{reutlinger2018extending}, the connection to causality is the main factor that underlies the explanatory force of CEs in XAI. We can see a textual CE in XAI as a true counterfactual conditional in which the antecedent describes a change in input features and the consequent a corresponding change in the classification. 

Research on CEs in Psychology concerning human-to-human interaction is another root and inspiration of the discussion in XAI \citep{byrne2016counterfactual,miller2019explanation}. Humans use CEs in their daily life when they explain behaviour or phenomena to each other, often in the form of a contrastive explanation highlighting the differences to the real scenario.  \cite{byrne2019counterfactuals} summarized the central findings on CEs in Psychology and evaluates their relevance to XAI. She points out that people tend to create CEs that: add information rather than delete, show better rather than worse outcomes, identify relevant cause-effect relationships, and change antecedents that are exceptional, controllable, action-based, recent, and not highly improbable.

The transfer of Lewis's account to generate explanations for the decisions of ML algorithms was first proposed by \cite{wachter2017counterfactual} who also drew the connection to the philosophical/psychological tradition of CEs. They argue that CEs have three intuitive functions: \emph{raise understanding}, \emph{give guidance for future actions}, and \emph{allow to contest decisions}.\footnote{It is not necessarily the case that all of these functions are or can be satisfied by one CE \citep{RussellCoherent}.} Also, they highlighted the legal relevance of CEs and argued that they satisfy the requirements proposed in the so-called 'right to explanation' as it is defined in Recital 71 of the European General Data Protection Regulation (GDPR). This law guarantees European citizens the right to obtain an explanation in cases they are subject to the fully automated decision-making of an algorithm \citep{voigt2017eu}.

\paragraph{History of AEs:}
Adversarial Examples have a less rich philosophical tradition, but instead a strong history in the robustness and reliability literature in computer science \citep{joseph2018adversarial}. \cite{fernandez2005model} describes robustness as ``the ability of a software to keep an 'acceptable' behavior...in spite of exceptional or unforeseen execution conditions.'' The reliability and robustness of computer systems have always been major concerns, especially in safety-critical applications such as health or the military sector. Critical elements can be the human interactors, hardware (e.g. sensors, hard drives, or processors), and the software. All kind of software is prone to erroneous behaviour \citep{kizza2013guide}, however, adversarial ML focuses particularly on the robustness of ML software.

For classical 'rule-based' software, the robustness can often be tested by formal verification \citep{d2008survey}. This becomes more difficult if systems interact dynamically with their environment or learn from data. Statistical Learning Theory tries to extend the idea of formal verification to statistical learning methods and gives theoretical guarantees for the performance of specific model-classes \citep{vapnik2013nature}. Unfortunately, good guarantees become unattainable for very broad and powerful model-classes such as for Deep Neural Networks and learning procedures like Stochastic Gradient Descent \citep{goodfellow2016deep}. What is special about the robustness of complex ML algorithms compared to others is that they are vulnerable to attacks even if common errors in model-selection have been avoided  \citep{bishop2006pattern,claeskens2008model,good2012common}. Moreover, the kind of attacks they are vulnerable to is highly unexpected and even put into question whether they learn anything meaningful at all \citep{szegedy}. The study of adversarial ML is not restricted to Deep Learning but also applies to classical ML models e.g. logistic regression \citep{dalvi2004adversarial}. 

The research in adversarial ML focuses on attacks on ML models by manipulated inputs and the defenses against such attacks. An AE describes an input to a model that is deliberately designed to effectively "fool" the model into misclassifying\footnote{From now on, we will mainly talk about misclassification and classifying. However, this is only to simplify our language usage. AEs are not restricted to classification tasks but also work on regression problems.} it. AEs occur even for ML algorithms with strong performances in testing-conditions. Since the changes from the original to the adversarial input are mostly \emph{imperceptible to humans}, AEs have been compared to optical illusions tailored to ML models \citep{elsayed2018adversarial}.

\cite{szegedy} and \cite{goodfellowEx} contributed milestones in the literature on AEs by not only providing ways to generate AEs but also attempting to explain their existence. \cite{szegedy} argued that AEs live mainly in spaces of low probability in the data-manifold. Therefore, they do not appear in either the training or the test dataset. Hence, artificial neural networks (ANNs) can have a low generalization error despite the existence of AEs. \cite{goodfellowEx} refuse this thesis and argue that AEs arise instead due to the linearity of many ML models including ANNs with semi-linear activations. \cite{tanay2016boundary} disagree and show that linearity is neither sufficient nor necessary to explain AEs. Instead, they claim that AEs lie slightly outside the real-data distribution close to tilted decision boundaries. They argue that the decision boundary is continuous outside the data-manifold and can therefore easily be crossed by AEs. A radically different view is proposed by \cite{ilyas2019adversarial} who show that AEs arise from highly predictive but non-robust features present in the training data. Hence, AEs are a human-centered phenomenon, the ML models, however, just rely on useful information in the data humans do not use.\footnote{Since it is extremely controversial why AEs exist, it is also hard to defend a system against them. It is even difficult to formulate the desired property an ML model should have concerning AEs \citep{bastani2016measuring,biggio2018wild}. Classical verification methods have to be modified because they explode computationally in the high-dimensional input spaces we are dealing with in ML. Since defense techniques are not relevant for CEs, we will not discuss them in the present paper. We advise the interested reader to \cite{serban2020adversarial}.}

\subsection{Role in ML}
\label{subsec:role}
 Due to the theoretical foundation, the practical applicability, and legal significance, the CE approach was quickly adopted by the XAI community as one method to explain individual predictions of ML models to end-users \citep{verma2020counterfactual}. Nevertheless, the method remains controversial and has often been accused of giving misleading explanations \citep{laugelDangersjournal, barocas,paez2019pragmatic}.

The trust we have in AI systems is and will be closely linked to the extent to which adversarial attacks are possible \citep{toreini2020relationship}. On the negative side, AEs can cause severe damage and security threats \citep{eykholt2018robust}. On the positive side, AEs can help us understand how the algorithm works \citep{ignatiev2019relating,tomsett2018failure} and therefore to understand what it has actually learned \citep{lu2017safetynet}. AEs can even concretely improve models \citep{bekoulis2018adversarial,stutz2019confidencecalibrated}.

Both CEs and AEs, play a great role in the ML landscape, namely for the trust people have in ML \citep{shin2021effects,toreini2020relationship}. CEs and AEs contribute to improving model understanding, identifying biases, and even offer methods to eliminate these biases through adversarial/counterfactual-training \citep{bekoulis2018adversarial,Sharma_2020}. However, while improving understanding and highlighting algorithmic problems is usually only a byproduct of AEs, it is the focus of CEs.  The deception of a system, on the other hand, is essential for AEs, but a potential byproduct of CEs in cases where they disclose too much information about the algorithm \citep{SokolFlach}.

\subsection{The Relation Between CEs and AEs}
\label{subsec:rel}
As mentioned in \Cref{sec:intro}, CEs and AEs derive from solutions to the same optimization problem \ref{eq:optimization}. While the close mathematical relationship between CEs and AEs has been frequently pointed out, their exact relationship remains controversial and there are a variety of opinions on the matter we present here in more detail. 

In one of the early papers on CEs, \cite{wachter2017counterfactual} note that an AE can be described as ``a counterfactual  by  a  different  name'' \citep[p.852]{wachter2017counterfactual}. They see one difference between counterfactuals and adversarials in the applied notion of distance arising from the misaligned aims, e.g. sparsity vs imperceptibility. The other difference they argue for is that while counterfactuals ought to describe closest possible worlds, AEs often result from 'impossible worlds' in the Lewisian sense i.e. unrealistic data-points. Additionally, they hint at methodological synergies between the two approaches, especially with respect to optimization techniques.

\cite{browne2020semantics} reject the two difference makers between CEs and AEs highlighted by \cite{wachter2017counterfactual} (distance metrics, possibility of worlds) as not definitional. They argue that using the ``wrong'' notion of distance may favor less relevant counterfactuals, but these are still ultimately potential explanations. Moreover, they reject the claim that adversarials must describe impossible worlds by pointing out that adversarial attacks can be carried out in real-world settings. Instead, they view counterfactuals and adversarials as formally equivalent. They argue that the key difference between CEs and AEs is not mathematical, but relies on the semantic properties of the input spaces. They point out that: ``Mathematically speaking, there is no difference between a vector of pixel values and a vector of semantically rich features.'' \citep[p.6]{browne2020semantics} They highlight the role of semantics in human-to-human explanation and claim that this difference makes CEs for image-data adversarials as AEs cannot be conveyed to an explainee in human-understandable terms.

\cite{verma2020counterfactual} see the terms CE and AE as non-interchangeable due to the different desiderata they must account for. They highlight tensions between the adversarial desideratum of imperceptibility and counterfactual desiderata like sparsity, closeness to the data-manifold, and actionability. According to \cite{grath2018interpretable} CEs and AEs are similar as both are example-based approaches. They describe the distinction between CEs and AEs as the difference between flipping and explaining decisions. They remark that CEs inform about the changes, while AEs aim at hiding those. \cite{laugel2019unjustified} agree that the two concepts show strong mathematical similarities. However, they also point to the difference in purpose and application. They note that CEs are mainly considered in the context of low-dimensional tabular data scenarios, whereas AEs are considered in less-structured domains like image/audio data. \cite{dandlmulti} and \cite{molnar2019} describe AEs as special CEs with the aim of deception. \cite{SokolFlach} discuss CEs in the context of AI safety. They make the case that CEs can disclose too much information about the model and thereby lead to AEs.

\section{Defining Concepts}
\label{sec:basicDef}
Evidently, there is a controversy around the relationship between CEs and AEs. Now, we will first resolve the conceptual controversy and building on that, we will present our definitions of CEs, AEs, and related terms.

\subsection{Conceptual Discussion}
\label{subsec:conDis}
In this section, we will discuss the accounts presented in \Cref{subsec:rel} and reflect their validity. We will propose that the relation of the counterfactual/adversarial to the true label and the proximity to the original data-point present the definitional distinction between CEs and AEs. In our arguments, we assume that the reader is familiar with the ideas behind \emph{decision boundaries, data manifolds, meaningless/unrealistic/unseen inputs, and distance metrics}. For readers who are not familiar with these concepts, we have provided a short glossary in \Cref{ap:glossary} where we explain these concepts with an illustrative example.

\paragraph{Two Names for the Same Objects:} Drawing from the fact that the same optimization problem can find CEs and AEs, \cite{wachter2017counterfactual} and \cite{browne2020semantics} conclude that they are the same mathematical objects. To evaluate this claim, imagine a model, e.g. an image classifier that, for all inputs for which a ground truth exists, assigns exactly this ground truth. Now, consider a particular prediction of this perfect algorithm. Via solving the optimization problem in \Cref{eq:optimization} we can generate counterfactuals. The CEs would be pointing to another input that receives a different assignment e.g. instead of the original image of a $3$, it shows a $9$ looking similar to that $3$. However, the system cannot be fooled by a modified image because it is always correct. Therefore, no AEs exist in that case and none of the generated counterfactuals is an AE. The case of a perfect algorithm shows that there are models for which we can reasonably generate CEs but no AEs. Consequently, they cannot generally be the same objects with different names. This shows that while there may be some cases where a vector can be called both counterfactual and adversarial, there must be a definitional difference between the two concepts.

\paragraph{The Two Differ in Aims:} \cite{verma2020counterfactual} point out that the terms are not interchangeable because ``while the optimization problem is similar to the one posed in counterfactual-generation, the desiderata are different.'' \citep[p.4]{verma2020counterfactual} By desiderata they mean additional requirements that are enforced on adversarials like imperceptibility or on counterfactuals like sparsity, closeness to the data-manifold and feasibility (For further explanations on the notions of sparsity, imperceptibility, closeness to the data-manifold, or feasibility see \Cref{sec:Generation}.). Usually, these different desiderata are realized in the different distance metrics applied. Therefore, the difference in aims corresponds to what \cite{wachter2017counterfactual} mean by claiming that AEs are not making use of appropriate distance metrics.

We agree with \cite{browne2020semantics} that the applied distances do not indicate a  definitional difference between CEs/AEs. We contend that whether the desiderata overlap or not, depends on the respective aims the user has with a CE/AE. Agents might also be interested in generating CEs to get guidance on how to deceive the system \citep{SokolFlach}. In such cases, imperceptibility will indeed be relevant, while sparsity or closeness to the data-manifold will be less relevant. Moreover, attackers could be interested in creating realistic AEs because they are more imperceptible. In such scenarios, closeness to the data-manifold or feasibility constraints are desirable properties of AEs. Also, both CEs and AEs can be relevant to better understand the model at hand and to improve it. 

If the desiderata are similar, so is the mathematical approach is. In such scenarios, good counterfactuals and adversarials may actually align and describe the same objects. However, a proper definitional distinction between concepts should be universal, objective, and independent of the agent's intentions. It requires necessary (and sufficient) criteria that make an object an instantiation of one object-class rather than another. The various desiderata are insufficient to account for differences between counterfactuals and adversarials in this strong sense.

\paragraph{Flipping and Explaining:}
\cite{grath2018interpretable} draws the distinction between CEs and AEs as the difference between explaining and flipping a decision. While the former point to changes in a meaningful way, the latter tries to hide those. We think that this is a solid observation, however, it shows a difference in presentation and not in definition. If the presentation style would be the whole difference, we would agree that they could mathematically be described as the same objects by a different name.

\paragraph{Low vs High-Dimensional Use Cases:} \cite{laugel2019unjustified}, \cite{wachter2017counterfactual}, and \cite{browne2020semantics} highlight the difference in use-cases. They argue, that while for CEs mainly low-dimensional and semantically meaningful features are used, AEs are mostly considered for high-dimensional image data with little semantic meaning of individual features. Therefore, the difference is not a difference of mathematical objects but rather a difference of semantic structure provided to generate an explanation/attack. In that sense, an AE is a CE that points to semantically non-interpretable factors. 

However, as discussed in \Cref{sec:useCases} the use-cases are increasingly overlapping. So, if \cite{browne2020semantics} would be right that the provided semantics in the input spaces is the crucial difference, authors studying AEs in low-dimensional setups would just directly use the approaches from the CE literature instead of developing new methods. According to their argumentation, the two approaches should be equivalent for low-dimensional setups. But, what we can notice is that e.g. \cite{ballet2019imperceptible} uses expert knowledge to generate imperceptible AEs for structured data by asking for features they find irrelevant for the decision at hand. Moreover, \cite{goyal2019,poyiadzi2020face} manage to give, as it seems, meaningful CEs also for high-dimensional input spaces without making use of higher-level semantic concepts the model creates while \cite{browne2020semantics} thought this is inevitable. These examples show that the semantic structure of the input space cannot account for a definitional distinction. Nevertheless, we agree that the difference between CEs and AEs is semantic in nature.

\paragraph{Misclassification:}
One obvious distinction that has largely been overseen by researchers is that adversarials must be necessarily misclassified while counterfactuals are agnostic in that respect. A correctly classified counterfactual is acceptable and often even desirable. On the other hand, if an adversarial were correctly classified, no one would call it an adversarial as it would provide no means to attack a target system. Consequently, misclassification is a necessary condition that any object called an adversarial must meet. This is different to the desiderata discussed above, which depend only on the goals of the agent with a CE or AE. Misclassification as a definitional distinction has been overseen since CEs and AEs can be generated by solving the same optimization problem \ref{eq:optimization}. How can it be that the same optimization problem is used to generate CEs for tabular-data models and AEs for image-data models? This is the crucial question that has to be assessed. It is strongly connected to the riddle the existence of AEs poses as discussed in \Cref{subsec:history}. Our analysis bases on the ideas of \cite{szegedy} and \cite{tanay2016boundary}.

We must look at image-classification models to answer why solutions to \Cref{eq:optimization} are mostly misclassified in that scenario. Complex image classifiers perform reasonably well on training data and highly similar inputs. In ``unseen regions'', on the other hand, they have to extrapolate and therefore perform worse. Since the input space is incredibly high-dimensional, the training data and therefore the data-manifold the algorithm approximates is comparably tiny. That means, there are many more meaningless, unrealistic, and unseen inputs than there are points in the training-data. The assignment of these inputs is not trustworthy and does not necessarily match the assignment of other nearby inputs. At the same time, there is usually a strongly limited number of classes that inputs are assigned to. Moreover, the training-data assigned to different classes have great distances. Hence, if we search for an input from another class but close to a given input, the probability is high that it is an input the algorithm has not seen, is unrealistic, or is meaningless and therefore where the algorithm is not reliable. Thus, the model will with high-probability misclassify this input. Often these close inputs are neither unrealistic nor meaningless as thought of by \cite{wachter2017counterfactual} but realistic. Completely unrealistic or meaningless inputs lie in greater distance to the original input. Realistic but unseen data-points make the dangerous AEs.

This explains why misclassified adversarials are generated in input spaces with high-dimensionality and little structure. The effect is even stronger if distances are applied that do not reflect what humans consider to be close inputs in the high-dimensional case. Minimal changes according to conceptually less-justified distances break the dependencies between variables present in the real world and therefore search for inputs in regions with less training-data support. This line of thought might suggest that the main reason why mostly adversarials are obtained by \Cref{eq:optimization} for image-classification is the use of distance metrics with little conceptual justification. Whether the right distance metric would yield fewer adversarials is, in our opinion, an empirical question that we cannot settle here. However, we will present our thoughts on this in \Cref{subsec:limitations}.

There are several reasons why counterfactuals generated in structured, low-dimensional input spaces are not generally adversarials. First, the models are often more robust and extrapolate better in unseen regions, also because background knowledge can more easily enter the model. Second, the real-world variables have a much simpler dependence structure compared to the high-dimensional image-data case. As additionally distances are picked that prefer sparse instead of distributed changes, these dependencies are often kept intact by the manipulations on the input vectors. Third, often additional constraints are added that make sure that the generated input stays close/within the data-manifold i.e. in regions where the model performs well (Further discussions of these constraints can be found in \Cref{subsec:Distances}).

Summed up, both counterfactuals and adversarials can be generated using the same method. However, that does not entail that they describe the same object class. Counterfactuals are agnostic with respect to the true label, whereas adversarials must be misclassified. From this perspective, counterfactuals could be considered the more general object-class. However, this conclusion would be drawn too early, since there is a second definitional difference, which we discuss next.

\paragraph{Proximity to the Original Input:}
Additionally to misclassification, we want to highlight a second, minor distinction between counterfactuals and adversarials, which is their tolerance with respect to proximity to the original input.

Closeness to the original input is usually a benefit for adversarials to make them less perceptible. However, an adversarial can still be used to attack a system if it is a little bit more distal to $x$ than another adversarial \citep{goodfellowEx}. Dependent on the aim of the attacker, this might even be desirable. Adversarials with greater distance to the decision boundary transfer better between different models, are often more effective, or more meaningful \citep{zhang2019limitations,elsayed2018adversarial}. 

For counterfactuals on the other side, closeness to the original input plays a significant role for the causal interpretation as discussed in \Cref{subsec:history}. Without maximal closeness, a counterfactual shows only a sufficient scenario for a different classification but not a necessary one. For example, assume we are in the loan-application setting from \Cref{sec:useCases}, where one point describes a maximally close counterfactual and the other a relatively close alternative input to $x$, both assigned to the same class. Assume moreover that the only difference between them is a change in gender from female to male. Then, even though such a change in gender would not impact the model-prediction, it would appear as a cause for the explainee receiving the alternative input. Such alternative inputs are less valuable than actual counterfactuals not only to data-subjects but also for model-developers examining the model. Thus, accepting 'close enough' but not maximally close inputs with a different classification as counterfactuals means either ignoring better CEs or admitting that the used distance is not perfectly adjusted for relevance in the given context.

Despite that difference in their tolerance with regards to proximity, we do not see this difference as equally essential as misclassification. If closeness is handled more loosely to generate ``CEs'', we might not gain real CEs, but still possibly relevant explanations. Thus, we do not entirely leave the category of objects. If on the other side we generate correctly classified inputs, we left the realm of attacks.

\subsection{Definitions}
\label{subsec:definitions}
Inspired by our discussion, we will lay out our basic definitions of CEs and AEs. These definitions will be also grounded in definitions well known in the literature (e.g. \cite{verma2020counterfactual,stepin} for CEs and \cite{szegedy,serban2020adversarial} for AEs), but extend them to fill conceptual gaps. We try to be maximally inclusive with respect to the usage of the terms in the general literature. However, due to the great number of papers on both fields \citep{yuan2019adversarial,verma2020counterfactual,serban2020adversarial,stepin} our framework will probably not be able to cover all usages.

Before we can define CEs and AEs, we need to know what we aim to explain or attack, namely machine learning models or the processes in which they are employed. We will restrict ourselves here to the highly common supervised learning setup. Moreover, we will focus on classification tasks. These restrictions have mainly the purpose to keep the analysis accessible. Many notions can be easily extended to other learning-paradigms.

\paragraph{Machine Learning Algorithms and Models:}
Assume we consider the relation of variables $\mathcal{X}:=\mathcal{X}_1\times\cdots\times \mathcal{X}_n$ and a (often one-dimensional) variable $\mathcal{Y}$. We can see these variables as random variables standing in a causal relation to each other. Let $X$ and $Y$ denote the co-domain of $\mathcal{X}$ respectively $\mathcal{Y}$.
A (supervised) \emph{machine learning algorithm} $\Phi$ is a procedure that based on a set of models $\mathcal{M}$, a labeled training dataset $\mathcal{D}_{Tr}:=\lbrace (x^1,y^1),\dots,(x^n,y^n)\rbrace$ with $n\in\N$, some hyperparameters $\mathcal{H}$, an optimization method $\mathcal{O}$, and a loss function $\mathcal{L}$ outputs a model $f\in\mathcal{M}$. This procedure $\Phi$ intuitively speaking searches for a model $f$ in the set $\mathcal{M}$, using method $\mathcal{O}$ and hyperparameters $\mathcal{H}$, that has a low prediction loss $\mathcal{L}$ on the training dataset $\mathcal{D}_{Tr}$. 

The model $f\in\mathcal{M}$ that is obtained by running the procedure $\Phi$ on a given input is called the \emph{machine learning model}. It can be described as a function $f:X\rightarrow Y$. This model ideally has a low bias measured by the loss function on the training dataset $\mathcal{D}_{Tr}$ and, moreover, a low generalization error on an unseen test dataset $\mathcal{D}_{Te}:=\lbrace (x^{n+1},y^{n+1}),\dots,(x^l,y^l)$ with $l>n$. That means that $f$ does predict values of $\mathcal{Y}$ from $\mathcal{X}$ in cases it has seen the correct assignment, but also for cases that have not been part of the training dataset $\mathcal{D}_{Tr}$. 

\paragraph{Counterfactuals and Adversarials:}
Unlike other authors, we distinguish between the mathematical objects that induce a CE/AE and the explanations/examples themselves. First, we will define the mathematical objects. For all the following definitions, assume we consider a fixed ML model $f$, a particular vector\footnote{This vector $x$ describes mostly a real-data instance.} $x\in X$ that is mapped by $f$ to a value $f(x)\in Y$, and a semi-metric
\footnote{A semi-metric on a space $X$ is a function $d:X\times X\rightarrow \R$ such that for all $x,x'\in X$ $d(x,x')\geq 0$, $d(x,x')=0 \Leftrightarrow x=x'$, and $d(x,x')=d(x',x)$.} $d(\cdot,\cdot)$ on space $X$.

\begin{defn}We call $x'\in X$ an \emph{alternative} to $x$ if $f(x')\neq f(x)$.
\end{defn}
In simple terms, $x'$ is an alternative to $x$ if it gets a different assignment by $f$.
\begin{defn}Let $\epsilon>0$. We call $x'_{\epsilon}$ an \emph{$\epsilon$-alternative} to $x$ if
\begin{equation*}
    d(x'_{\epsilon}, x)<\epsilon \text{ and }x'_{\epsilon} \text{ is an alternative to }x.
\end{equation*}
\end{defn}
We can think of $x'_{\epsilon}$ as a step away from $x$ for which we cross a decision boundary of the model but stay within a local $\epsilon$-environment around $x$.
\begin{defn}We call $c_{x}\in X$ a \emph{counterfactual} to $x$ if
\begin{equation*}
    d(c_{x}, x) \text{ minimal subject to } f(c_{x})\neq f(x).
\end{equation*}
\end{defn}
Staying in the narrative, a counterfactual describes the shortest\footnote{with respect to $d(\cdot,\cdot)$} step that crosses a decision boundary. Notice that this closest vector does not have to be unique, there might exist a variety of vectors in equal distance. 

A \emph{true label} $y_{x',true}\in Y$ for a vector $x'\in X$ describes the objectively correct label that the input-vector $x'$ should be assigned to. This ground-truth is often given by expert human evaluation. Not for all inputs there exists such a true label. The reason might be that the correct assignment is controversial even among expert evaluators  or the considered input is unrealistic. Why are such unrealistic inputs relevant? As introduced above, image$(\mathcal{X})\subseteq X$. That means that in cases where the subset-relation is strict, our model $f$ is defined on data-points that do not realistically occur in the real world.

\begin{defn}
We call a vector $x'\in X$ \emph{misclassified} if $f(x')\neq y_{x',true}$.
\end{defn}
A misclassification describes a mistake made by the algorithm relative to an expert-human assignment.

\begin{defn}Let $\epsilon>0$. We call $a_{x,\epsilon}\in X$ an \emph{adversarial} to $x$ if 
\begin{equation*}
    a_{x,\epsilon} \text{is an }\epsilon\text{-alternative and misclassified}.
\end{equation*}
\end{defn}

In the literature, no clear definitional distinction is drawn between counterfactuals and adversarials. However, as we have argued in \Cref{subsec:conDis}, we believe that the distinctions we have introduced are conceptually necessary. The definitions of counterfactuals and adversarials differ in two aspects: the relation to the true instance label and the constraint of how close the respective data-point must be. The misclassification of adversarials enters the definition by enforcing it as an additional necessary condition. Note that this entails that only inputs for which a ground-truth exists can in our definition be called adversarials.

The second definitional difference we introduce is that counterfactuals must be maximally close data-points, while adversarials need only be within an $\epsilon-$environment around the original input $x$. This relaxed condition on adversarials is introduced via defining them as $\epsilon-$alternatives. This means, whether an input is called an adversarial or not depends on how close the attacker requires the input to be. If the constraint is put too strong i.e. if $\epsilon$ is too small, there might not exist any adversarials within that environment. If, on the other side, the constraint is set very high and similarity to the original input is not of great importance, even inputs that rather dissimilar to the original input can count as proper adversarials. Unlike adversarials, counterfactuals always exist as long as there exists an alternative to $x$, plus only maximally close alternatives count as proper counterfactuals.

Especially counterfactuals, but also adversarials are often \emph{targeted}. That means, the generated vector should not only belong to a different class than the original vector but to a specific desired class. For counterfactuals, this may be from the perspective of the end-user who wants to get her loan application accepted or the model-engineer who wants to check whether the model can distinguish an input from other inputs of a specific object-class (See \Cref{sec:useCases}). For adversarials, this may be the desired classification from the perspective of the attacker of the system (E.g. Whatever is next to this sticker is a toaster \citep{brown2017adversarial}.). In such cases we talk about \emph{targeted ($\epsilon$)-counterfactuals/adversarials}. More formally, let $y_{des}\in Y$ with $f(x)\neq y_{des}$ denote the desired outcome of a stakeholder given such a desired outcome exists.
\begin{defn}We call $x'\in X$ \emph{$y_{des}$-targeted} if $f(x')=y_{des}$.
\end{defn}

\paragraph{CEs and AEs:}So far, we have only discussed vectors living in a space $X$. How do we get from these vectors to explanations or attacks? 
\begin{defn}$ $

\begin{itemize}
    \item A \emph{contrastive explanation (CON)} is a presentation of an alternative $x'$ in contrast to $x$ understandable to a human agent.
    \item A \emph{counterfactual explanation (CE)} is a presentation of a counterfactual $c_{x}$ understandable to a human agent.
    \item  An \emph{adversarial example (AE)} is the depiction of an adversarial $a_x$.
\end{itemize}
\end{defn}
Notice that while every counterfactual and every adversarial describes an alternative, not every CE or AE is a CON. CONs must be presented as a contrast between $x'$ and $x$. Possible presentation styles for CEs/AEs include the presentation in form of an (English-)conditional of type III for tabular data, an image for visual-data, or a sound for auditory-data. For tabular data, we use the property that the input features in such scenarios are interpretable. That means they have semantic meaning and can be expressed by human language concepts.

Assume we are in such a tabular-data scenario where $x=(x_1,\dots,x_n)$ describes the original vector and $c_x=(c_{x_1},\dots,c_{x_n})$ one of its targeted-counterfactuals. Now, consider the vector $c_x-x$. $p\leq n$ of this vector's values will be non zero. Assume $k_1,\dots,k_p$ describe the names of these non-zero entries of the vector and $e_{k_1},\dots,e_{k_p}$ their respective values. The (contrastive) CE in this scenario would be:
\begin{itemize}
    \item[] If P had a $e_{k_1},\dots,e_{k_p}$ higher/lower value in $k_1,\dots,k_p$, she would have reached her desired classification instead of $f(x)$.
\end{itemize}

For image-data, we can use the fact that vectors in such spaces can be visualized directly in their image representation. Examples have been shown both for CEs and AEs in \Cref{sec:useCases}. The same holds for auditory data-inputs which can be presented as a sound.

As mentioned above, often there is not one unique counterfactual to a given vector $x$. Therefore, there is not one unique correct CE. Worse, often different CEs are incompatible. We call the fact that there are several equally ``good'' explanations for the same prediction the \emph{Rashomon effect} \citep{molnar2019}. Several ways to deal with this problem have been proposed. \cite{mothilal2019explaining}, \cite{moore2019explaining}, \cite{wachter2017counterfactual}, and \cite{dandlmulti} propose to present various CEs dependent on the specific aim of a user. However, then the question arises, how many and which ones?  Others propose to select a single CE according to relevance \citep{fernandezloria2020explaining} or a quality standard set by the user, such as complexity \citep{SokolFlach}. We think the question the Rashomon effect poses is still open to debate. AEs are unique neither. However, as AEs must not cohere, nor be necessarily presented to humans, this plays no role.

\paragraph{Model-Level and Real-World:} One distinction that is often overlooked is the difference between an explanation/attack on the model-level and the real-world. We need to be clear about whether we want to explain/attack the model or the modeled process. Generally, the former is much easier to accomplish than the latter. We can only move from a model explanation/attack to a process explanation/attack if the model itself, and also the translation of our inputs, preserve the essential structure of the process \citep{molnar2020pitfalls}. There are two scenarios for which the distinction between the two levels is relevant: it is relevant for CEs if a user is interested in recourse to attain a desired outcome \citep{KarSchVal20}. It is relevant for AEs if an attacker aims to deceive an ML system deployed in the physical world \citep{kurakin2016adversarial}.

To give two examples that highlight the difference between model-level and real-world explanations/attacks, we reconsider the examples from \Cref{sec:useCases}. The presented CE in the loan application setting was: ``If P had a $5,000$ \euro\, p.a. higher salary and an outstanding loan less, her loan application would have been accepted.'' This explanation clearly tells us something about the employed model, namely about the assignment for a particular alternative. However, P could take this as an action recommendation in the sense that if she raises her salary and paid her outstanding loan, she will receive the loan she applied for. Unfortunately, things are not that simple in the real world. P has to work hard to raise her salary and pay her open loan, this does not happen in zero time. By the time she reaches the required threshold, she may be five years older and her loan application will be rejected again, this time due to her advanced age or because a different algorithm is now used \citep{venkatasubramanian2020philosophical}. So the transfer from the model explanation to an action recommendation for recourse is not as easy.

A similar example can be shown for AEs. Consider the Hand-Written Digits Recognition Scenario from \Cref{sec:useCases} where an attacker aims to money-pump the postal service. The AEs presented are clearly inputs that trick the model. However, if she now aims to make the step to a real-world fraud, she has to print them out. A bad printing, different colors, alternative background, changed angles, or the camera employed by the postal service will impact which input the model receives. Thus, the AE might not work in the postal-service hand-written digits recognition service but only in the artificial setting where we can directly manipulate the input the model receives.

For both CEs and AEs, we need to know the employment context and the required functionality in order to be clear about what level we are dealing with. The work of \citep{KarSchVal20,mahajan-2019-preserving-constraints-counterfactual-explanations} on algorithmic recourse and the work of \cite{kurakin2016adversarial,lu2017no,athalye2018synthesizing} on adversarial examples in the physical world have alerted the CE and AE communities to the importance of the two different levels. The two levels collapse only for artificial settings in which the model perfectly matches the process \citep{KarSchVal20}.

\begin{defn}
We say a CE/AE operates at the \emph{real-world level} if it describes changes in $\mathcal{X}$ that result in changes in $\mathcal{Y}$. We say that a CE/AE operates at the \emph{model-level} if it describes changes in $X$ that result in changes in $Y$.
\end{defn}

\section{Generation of CEs/AEs}
\label{sec:Generation}
So far we have motivated and discussed the formal definitions of CEs/AEs. Now, we move from the definition to their generation. Again, we will focus on the connections between the two fields. Before we start, it is important to note that the generation methods for AEs do generally not guarantee success i.e. it is unclear whether the generated input vector is misclassified. Instead, misclassification is particularly in image-classification still often reached accidentally as discussed in \Cref{subsec:conDis}.

\subsection{General Approaches}
\label{subsec:GenSolApproaches}
\paragraph{Optimization Problem:}The most common approach to find CEs/AEs is to formulate and solve an optimization problem. Such a problem formulation is already present in the definition of CEs/AEs, however, this is optimization under side conditions and therefore not easy to solve. Instead, the standard formulation as a single objective optimization problem is \Cref{eq:optimization} that led to the confusion discussed in \Cref{subsec:conDis}.

For both CEs and AEs there exist many other formulations as an optimization problem. E.g. for CEs \cite{poyiadzi2020face}, \cite{kanamori2020dace}, and \cite{van2019interpretable} add additional terms to \Cref{eq:optimization} encoding further desiderata (See aims and distances below), \cite{dandlmulti} instead add these desiderata by formulating a multi-objective optimizations problem, and \cite{karimi2020model} formulate a search for the smallest intervention on the variables needed to attain a change in classification. Similarly to the former formulations for CEs, there exist approaches to AEs like \citep{carlini2017towards,moosavi2017universal} which modify the objective from \Cref{eq:optimization} to obtain desired properties like computational efficiency or universality of an AE. Other optimization problems also take into account transformations of background or objects and generate AEs whose classification is invariant under such transformations \citep{eykholt2018robust,brown2017adversarial,athalye2018synthesizing}.

\paragraph{Generative Networks:}A second way to generate CEs/AEs that has been fruitfully applied is the use of generative networks that generate CEs/AEs for a given input. This technique is widespread for both AEs \citep{goodfellow2014generative,zhao2017generating,yuan2019adversarial} and CEs \citep{mahajan-2019-preserving-constraints-counterfactual-explanations,van2019interpretable,pawelczyk2020learning}.

\paragraph{Sensitivity Analysis:}A third approach that is almost exclusively used by the AEs community is sensitivity analysis. Information about the gradient \citep{goodfellowEx,lyu2015unified} or Jacobian \citep{papernot2016limitations} of the function in the specific input is used to make a step in the direction of the decision boundary to a different class. \cite{moore2019explaining} is the only example we are aware of who use this approach to generate CEs. One reason why such approaches have probably not been picked up in the CE-literature is that it has limited conceptual justification e.g. with respect to minimal distance as we discuss in \Cref{sec:discussion}.

\subsection{Distances}
\label{subsec:Distances}
All approaches to generate AEs necessitate an underlying notion of distance, mainly for the inputs space but often also for the output space. Researchers worked with a high variety of distances. Often the distances encode specific desiderata researchers want CEs/AEs to satisfy. For both fields, the question for the right distance for a given use-case is considered an open problem \citep{serban2020adversarial,verma2020counterfactual}. Since every norm induces a metric, we will use the names of the norms and generally talk about distances.

\paragraph{Sparsity and Imperceptibility:}
Since explanations must be comprehensible to humans with limited temporal and cognitive resources, CEs should point to a few crucial features. Therefore, distances are preferred that take into account sparsity. For adversarials on the other side, a common aim is imperceptibility. Changes from the original input to the modified input should be hard to grasp for human observers. While these desiderata often lead to conflicting notions of distance, they also can coincide. E.g. the $L_0$ and $L_1$ norm have both been fruitfully been applied to generate sparse counterfactuals \citep{dandlmulti,wachter2017counterfactual} and imperceptible AEs \citep{Su_2019,tramer2019adversarial,pawelczyk2020learning}.

However, some distances to attain sparsity of counterfactuals have not been used to reach imperceptibility of AEs. One way by which sparsity can be guaranteed is to explicitly put a constraint on the number of features allowed to change \citep{kanamori2020dace,ustun2019actionable,SokolFlach}. Another is to constrain the number of actions that can be taken, but not the number of the corresponding feature changes \citep{KarSchVal20}. To attain imperceptibility of AEs, the distances are more diverse. Common examples are the $L_2$ \citep{moosavi2016deepfool} and $L_\infty$ \citep{goodfellowEx,elsayed2018adversarial} norm for distributed changes which often makes AEs look identical to the input they origin from. Other norms, more inspired by human perception are the Wasserstein-distance \citep{wong2019wasserstein}, using physical parameters underlying
the image formation process \citep{liu2018beyond}, or the Perceptual Adversarial Similarity Score \citep{rozsa2016adversarial}.

\paragraph{Plausibility and Misclassification:}
CEs should not present an entirely unrealistic alternative scenario to the explainee. Instead, the recommended alternative should be within reach and if possible it should be feasible for agents to perform actions based on these alternatives. This often means that the counterfactual lies in the natural data-distribution. AEs must by definition be misclassified, which as discussed in \Cref{subsec:conDis} is often easier to reach on the edges or slightly outside the natural data-manifold. We see an antagonism between the goal of realism of CEs and the misclassification of AEs from which the applied distances can potentially profit from.

One common way to attain plausibility is to take into account the distance of the CE to the closest training-datapoint \citep{kanamori2020dace,dandlmulti,Sharma_2020} or the allowed path to the counterfactual \citep{poyiadzi2020face}. Often, additional constraints are posed such that only actionable features should be changed to avoid non-helpful recommendations \citep{ustun2019actionable}. Another way is to take into account the causal structure of the real-world features. If a counterfactual arises realistically from an intervention on some of these features, the corresponding CE is plausible \citep{mahajan-2019-preserving-constraints-counterfactual-explanations,KarSchVal20}. 

To attain misclassified inputs, it is generally reasonable to search in low-probability areas of the data-manifold \citep{szegedy} or even outside of it \citep{tanay2016boundary}. Therefore, most distances for AEs do not respect the causal structure between the corresponding real-world variables. Some even act directly against the causal structure and modify only irrelevant features \citep{ballet2019imperceptible} or, just as for CEs, put constraints on the potential changes \citep{cartella2021adversarial}. Particularly noteworthy is the distance of \cite{moosavi2017universal} who encode the robustness of the flip in classification and also the work of \cite{carlini2017towards} who compare the misclassification between different applied distances. Interestingly, it has been found that a greater distance to the given decision boundary guarantees more robustness of misclassification why many do not search for minimal but only close adversarials \citep{zhang2019limitations}.

\paragraph{Contestability and Misclassification:}
CEs should allow explainees to detect adverse or wrong decisions. If the explainee is an end-user, this could be the case if she feels judged unfairly
 \citep{kusner2017counterfactual, asher2020adequate}. On the other side, if the explainee is the model-engineer, this could mean CEs reveal bugs. Again, AEs must be misclassified. Decision-making mistakes are the common denominator of the contestability reached by CEs and misclassification provided by AEs. Various ways have been proposed to encode these aims.
 
 \cite{RussellCoherent} provide contestability by presenting a range of diverse CEs in which different features were modified. This makes sure there will be some CEs that provide grounds to contest the decision. \cite{Sharma_2020} define protected properties like ethnicity and focus on changes in these features in their distance. \cite{laugel2019unjustified} discuss how standard norms like $L_1$ can lead to unjustified CEs since they arise from inputs outside the training-data. \cite{hashemi2020permuteattack} combines CEs and AEs to evaluate the weaknesses of a given model. They use both, the $L_0$ and $L_2$ norm plus focus on protected features in search for realistic but misclassified counterfactuals. In a similar vein, \cite{ballet2019imperceptible} assign importance weights to features and through these weights they define weighted $L_p$ norm where changes in more important features have a lower weight and are therefore more likely to change in the optimization process. \cite{cartella2021adversarial} extend their work and put additional constraints to keep the adversarials realistic but still fraudulent. Especially the last three examples show the great overlap between the goals of contestability and misclassification.
 
\subsection{Model-Access}
As we have discussed above, we do not need to define an optimization problem to generate counterfactuals or adversarials. However, different solution methods differ in the degree of model-access they need. We distinguish between black-box and white-box scenarios. In a black-box scenario, explainers/attackers can only query the model for some inputs they provide and receive the corresponding output. In a white-box scenario, the explainer/attacker has full model access. We can further distinguish between methods that only work for a particular model-class and methods that are model-agnostic. All black-box solvers work for any model. For white-box solvers, some only need access to gradients and therefore require a differentiable model and those that are specific to a particular model-class e.g. linear models but can therefore often handle mixed-data. Interestingly, even though white-box scenarios are more realistic for explainers and black-box scenarios more commonly occur for attackers, the literature shows tendencies in the opposite directions.\footnote{See \cite{serban2020adversarial} and \cite{verma2020counterfactual} who notice the respective tendencies in their surveys. They explain this by the chances to explore more in white box settings and the computational problems of black-box attacks in high-dimensional use cases (See \Cref{sec:useCases}).}

Many solvers rely on access to the models gradients e.g. for CEs \citep{wachter2017counterfactual,mothilal2019explaining,pawelczyk2020learning,mahajan-2019-preserving-constraints-counterfactual-explanations} or for AEs \citep{szegedy, athalye2018synthesizing, brown2017adversarial,ballet2019imperceptible}. Other solvers for CEs are model-specific and require full model-access such as mixed-integer linear program solvers \citep{ustun2019actionable, RussellCoherent,kanamori2020dace} or solvers tailored for decision trees \citep{tolomei2017interpretable}. For AEs some solvers require neural network feature representations \citep{SabourCFF15}. However, several solvers can deal with a black-box setup. Common in both literatures are evolutionary algorithms e.g. for CEs \citep{Sharma_2020, dandlmulti} and for AEs \citep{guo2019simple,alzantot2019genattack, Su_2019}. Very prominent for AEs are also the approximation of gradients by symmetric differences \citep{chen2017zoo} and the usage of surrogate models \citep{papernot2017practical}. Especially the latter approach is interesting as it is based on the transferability of AEs between different models optimized for the same task.

We see that many solvers are fruitfully used in both domains. It will be seen whether surrogate model-based approaches also find their way into the CE literature. We find the use of them for CEs conceptually controversial as the faithfulness to the model is more critical for an explanation than for an attack (Also see \Cref{sec:discussion} for a short discussion of this point.)).
\footnote{A first approach to use a surrogate model to generate similar explanations to CEs was proposed by \cite{guidotti2018local}.}

\section{Discussion}
\label{sec:discussion}

In this paper, we discussed the relationship between CEs and AEs. We argued that the definitional difference between the two object classes consists in their relation to the true data labels (i.e. adversarials must necessarily be misclassified) and their proximity to the original data-point (i.e. counterfactuals must be maximally close to the original input). Based on this, we introduced formal definitions for the key concepts of the fields. In addition, we have highlighted similarities and differences between the two fields in terms of use cases, solution methods, and distance metrics.

\subsection{Relevance} Our work adds a new viewpoint to the discussion of the relationship between CEs and AEs. Eventually, we hope that our work can form the basis for merging the two fields. Based on our arguments and the formal definitions inspired by them, adversarials can be seen as special cases of (more distal) misclassified counterfactuals. Especially when it comes to CEs for which misclassification is a desirable property, such as CEs for contesting adverse decisions, detecting bugs, or improving model-robustness, we see potential synergies. We believe that a solid conceptual discussion becomes more important as these functions of CEs are emphasized and as application domains overlap (e.g., AEs in lending, CEs for image classification).

Our work also has a clear practical relevance. The conceptual arguments for the maximal proximity of counterfactuals make clear that generating counterfactuals via sensitivity analysis, as proposed by \cite{moore2019explaining}, or using surrogate model approaches could be problematic. In the case of sensitivity analysis, maximal proximity to the original input is not guaranteed and hence the corresponding CEs have less explanatory power. Surrogate models might not be sufficiently faithful to the original model and therefore lead to bad/misleading explanations. As we discussed, solution methods to find CEs can also generate AEs, but the reverse can be problematic.

What we have shown in terms of the current literature is that there is a large amount of overlap. We have also suggested which parts are good candidates for transfers. However, as we have made clear, such transfers of mathematical frameworks or approaches require conceptual justification. While transferring gradient-based solution techniques from the AE literature to generate counterfactuals, as proposed by \cite{wachter2017counterfactual}, is conceptually unproblematic, using counterfactuals to measure the robustness of a model, as suggested by \cite{Sharma_2020}, will not work for tabular data scenarios.

\subsection{Limitations and Open Problems} 
\label{subsec:limitations}
\paragraph{Misclassification Formalized:}Our work points to an important weak spot of the current AE literature: misclassification is achieved more or less by accident in the image domain, but is not clearly formalized. Such a formalization of misclassification would gretly advance the merging process between CEs and AEs. It may be considered a limitation of our work that we have not provided this formalization but instead referred to the true data-labels, which are either expensive to obtain or simply unknown. Nevertheless, we want to provide a roadmap of what such a formalization might look like.

We believe that \cite{ballet2019imperceptible} made the first solid contributions to a formal representation without requiring the ground-truth data labels. In our opinion, a good candidate framework for generalizing their approach is causal modeling \citep{pearl2009causality}. If we have a true causal model, misclassification is obtained by modifying a correctly classified input sufficiently to change the classification, but in a way that violates the causal structure. We suggest that adversarials can be viewed as small modifications in causally irrelevant features that unjustly influence the prediction.

Unfortunately, approaching the problem of misclassification from a causal modeling perspective also comes with strong requirements: we need a structural causal model or at least a causal graph. Obtaining such models is extremely difficult \citep{pearl2009causality,scholkopf2019causality}, and when dealing with conceptually lower-order features such as pixels or sounds, causal models might even be the wrong descriptive language. Still, we think that even limited causal knowledge about, e.g. parts of the causal graph or some of the structural equation, might suffice in many contexts to prove that a change in classification is unjustified. Moreover, for conceptually less-structured feature spaces, higher-order causal models \citep{beckers2019abstracting} where features such as objects are supervened by lower-order features such as pixels may provide the right level of description to define misclassification.

\paragraph{Distances on Unstructured Spaces:}
In our discussion in \Cref{subsec:conDis} on misclassification, we gave reasons why most inputs that solve \Cref{eq:optimization} are misclassified. We argued that theoretically poorly justified distance metrics are one of the reasons for this phenomenon. However, we did not address whether this might be the only reason for this behavior and whether this would still be the case if we had conceptually well-justified distances on high dimensional spaces with little semantics such as pixel spaces.

We believe that this is an empirical question we could not settle in this paper. The standard way for approaching it would be to move the distances from raw features such as pixels to higher-order features such as object properties. It has often been pointed out that deep-learning algorithms based on convolutional neural networks (CNNs) \citep{goodfellow2016deep} automatically find semantically meaningful features in layers close to the output space \citep{zhang2018visual,bau2017network}. For example, one could define a distance function on the feature space just before the so-called dense layer in CNNs, which is responsible for classification. 

While we consider this is a promising direction for future research, there are good reasons to remain skeptical. First, unfortunately, it is not so easy to assign specific semantic meaning to these high-level features, since some of them are poly-semantic and  are triggered by quite different inputs \citep{olah2020zoom}. Distance measures on such features may therefore also be conceptually unjustified and the problem remains. Second, examples of AEs, such as those given \cite{szegedy} or \cite{goodfellowEx}, seem to show images that are almost identical to the original image. Hence, conceptually well-justified distance functions should also assign a low distance to these images, and consequently they will still be generated by solving \Cref{eq:optimization}. Following \citep{ilyas2019adversarial}, we think that AEs are generated by \Cref{eq:optimization} not only because we apply the wrong distance function, but also because the ML model has not really learned the robust concepts that humans use to distinguish objects.

\paragraph{Explanations and Deceptions:} We have not discussed the conceptual relationship between illusions and explanations more generally (e.g. the relation between everyday life explanations and cognitive biases or optical illusions), but have focused only on CEs/AEs in ML. In what sense can an illusion explain a phenomenon? How can an explanation lead to a deception? Is there an underlying conceptual or even cognitive connection between explaining and deceiving? We do find these questions, and the possible embedding of our CE/AE discussion within them, intriguing. For now, however, we leave these deep and difficult philosophical/psychological questions to other researchers.

\bibliography{library}
\newpage
\appendix
\section{Glossary}
\label{ap:glossary}
\begin{figure}[h]
\centering
  \includegraphics[width=0.7\linewidth]{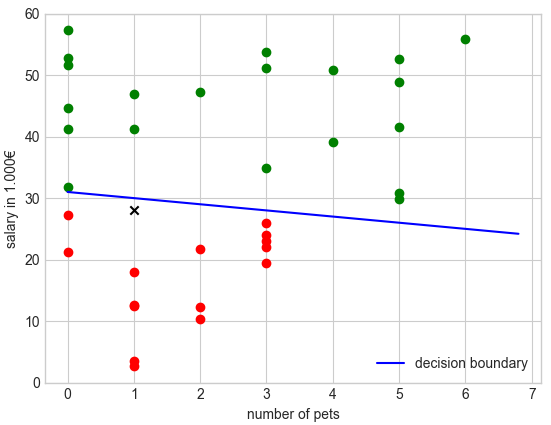}
  \caption{This figure depicts the decision behavior of a simple classifier. It describes the scenario from \Cref{sec:useCases}, which is inspired by \cite{ballet2019imperceptible}. The classifier uses two features, salary and number of pets, to decide whether to approve or reject a loan application. The green dots are the training data labeled as approved, the red dots are the training data labeled as rejected. The blue line describes the decision boundary of the classifier.}
  \label{fig:glossary}
\end{figure}

We will shortly explain the following terms with the help of the example depicted in \Cref{fig:glossary}. As in \Cref{subsec:definitions} we call $f:X\rightarrow Y$ the classifier, $X$ the input space, and $Y$ the output space.

\textbf{Decision boundary:} 
In the example, the decision boundary is described by the blue line. All inputs above the decision boundary are labeled ``approved'', all inputs below the blue line are labeled ``rejected''. Crossing the decision boundary means that a point is moved from one side of the decision boundary to the other. For example, the individual represented by the black ``x'' at position $(1,28)$ might cross the decision boundary by moving his salary up 1,000\euro\; or by buying two more pets.

More generally, we can describe a decision boundary as a hypersurface in space $Y$ that separates one class from another. These hypersurfaces are induced by the classification model $f:X\rightarrow Y$.

\textbf{Data-manifold:} 
In our example, the green and red points lie within the data-manifold of realistic data samples. However, there is no point number or negative number of pets, so such instances would lie outside the data-manifold.

More generally, a data-manifold describes a subset (often a hypersurface) of the spaces $X\times Y$ that arises naturally from a data-generating mechanism. A data-manifold encompasses the statistical population. The training and test data are usually a sample from this population.

\textbf{Meaningless, unrealistic, or unseen inputs:}
\begin{itemize}
    \item Meaningless: An example of a meaningless input in our scenario would be a person with a negative number of pets. It describes an input that makes no sense to us, but is contained in the space $X$.
    \item Unrealistic: An example of an unrealistic input in our scenario would be a person with five million pets. It describes an input we can understand, but that most likely does not occur in the real world.
    \item Unseen but realistic: An example of an unseen input in our scenario would be a person who earns 29,000\euro\; and has four pets. It describes an input that may realistically occur in the real world, but was not part of the training data.
\end{itemize}

\textbf{Conceptually (un-)justified distance metrics:}
Conceptually unjustified distance metrics assign small distances to inputs that are not similar from a conceptual standpoint. In our example, a distance function might assign a small distance to the points $x_1=(0,10)$ and $x_2=(22,10)$.  This would make $x_2$, which lies far outside the data manifold and is assigned to the ``approved'' class by the model, a potential counterfactual for $x_1$. However, $x_2$ is highly unrealistic as 20 pets are a lot and it breaks the dependence that 20 pets are probably too expensive for an income of 10,000\euro\; per year. This dependency problem is more severe for pixel spaces, since pixels have strong dependencies in the real world with their neighboring pixels. Moreover, an image in the form of a set of pixels represents an image of objects to humans, a fact that is difficult to account for with a metric.
\newpage

\section*{Declarations}
\subsection*{Funding}
This work was supported by the Graduate School of Systemic Neuroscience (GSN) of the LMU Munich.
\subsection*{Conflicts of interest/Competing interests} 
The author has no conflict of interest.
\subsection*{Availability of data and material}
Not Applicable
\subsection*{Code availability}
Not Applicable
\end{document}